\documentclass{article}

\usepackage[preprint]{neurips_2025}

\usepackage[utf8]{inputenc}
\usepackage[T1]{fontenc}
\usepackage{hyperref}
\usepackage{url}
\usepackage{booktabs}
\usepackage{amsfonts}
\usepackage{nicefrac}
\usepackage{microtype}
\usepackage{xcolor}

\usepackage{calc}
\usepackage{subcaption}
\usepackage{mathtools}
\usepackage{amsmath}
\usepackage{import}
\usepackage{bm}
\usepackage{tikz}
\usetikzlibrary{positioning, arrows.meta}
\usepackage{thmtools}
\usepackage{thm-restate}

\usepackage{xspace}
\usepackage{thmtools} 
\usepackage{thm-restate}
\usepackage{microtype}
\usepackage{graphicx}
\usepackage{booktabs}
\usepackage{multirow}

\usepackage{wrapfig}
\usepackage{multirow}
\usepackage[percent]{overpic}

\usepackage[capitalize,noabbrev]{cleveref}

\newcommand{\rpmjoint}[1]{\tilde{p}_{\theta}^{\bm{X}}\left( {#1} \right)}
\newcommand{\rpmf}[1]{f_{\phi_t}\left( {#1} \right)} 
\newcommand{\rpmF}[1]{F_{\phi_t}\left( {#1} \right)} 
\newcommand{\rpmpx}[1]{p_0\left( {#1} \right)} 
\newcommand{\rpmaux}[2]{g^{#1}_{#2}\lrify} 
\newcommand{\rpmfdelta}{f^\Delta_{\phi}\lrify} 

\newcommand{\subfigref}[2]{Figure~\hyperref[#1]{\ref*{#1}#2}}

\newcommand\blfootnote[1]{%
  \begin{NoHyper}
  \renewcommand\thefootnote{}\footnote{#1}%
  \addtocounter{footnote}{-1}%
  \end{NoHyper}
}

\newcommand{\rpmfj}[1]{f_{\phi_j}\left( {#1} \right)} 
\newcommand{\rpmFj}[1]{F_{\phi_j}\left( {#1} \right)} 

\newcommand\nn[1][n]{^{#1}}

\newcommand*\diff{\mathop{}\!\mathrm{d}}
\newcommand{\normal}[1]{\setN\left( {#1} \right)}
\newcommand{\ave}[2]{\left\langle {#1} \right\rangle_{#2}} 
\newcommand{\kl}[2]{\mathsf{KL}\left(#1\,\middle\lvert\middle\rvert\,#2\right)}
\newcommand{\ind}{\perp\!\!\!\!\perp}

\newcommand{\qed}{\hfill \ensuremath{\Box}}

\newcommand{\reals}{\mbox{\(\mathbb R\)}}

\newcommand{\setD}{{\mathcal{D}}}

\newcommand{\setF}{{\mathcal{F}}}
\newcommand{\setG}{{\mathcal{G}}}
\newcommand{\setH}{{\mathcal{H}}}

\newcommand{\setL}{{\mathcal{L}}}

\newcommand{\setN}{{\mathcal{N}}}

\newcommand{\setX}{{\mathcal{X}}}

\newcommand{\setZ}{{\mathcal{Z}}}

\newcommand{\model}{RP-GSSM\xspace}
\input{m.symbols}

\title{Maximum Likelihood Learning of Latent Dynamics Without Reconstruction}

\author{%
  Samo Hromadka\textsuperscript{1,$\ast$} \And Kai Biegun\textsuperscript{2} \And Lior Fox\textsuperscript{1} \And James Heald\textsuperscript{1} \And Maneesh Sahani\textsuperscript{1}
}

\begin{document}

\newgeometry{
textwidth=5.6in,
textheight=9.1in,
centering,
top=0.9in
}

\maketitle

\begin{abstract}
We introduce a novel unsupervised learning method for time series data with latent dynamical structure: the recognition-parametrized Gaussian state space model (\model).
The \model is a probabilistic model that learns Markovian Gaussian latents explaining statistical dependence between observations at different time steps, combining the intuition of contrastive methods with the flexible tools of probabilistic generative models.
Unlike contrastive approaches, the \model is a valid probabilistic model learned via maximum likelihood.
Unlike generative approaches, the \model has no need for an explicit network mapping from latents to observations, allowing it to focus model capacity on inference of latents.
The model is both tractable and expressive: it admits exact inference thanks to its jointly Gaussian latent prior, while maintaining expressivity with an arbitrarily nonlinear neural network link between observations and latents.
These qualities allow the \model to learn task-relevant latents without ad-hoc regularization, auxiliary losses, or optimizer scheduling.
We show how this approach outperforms alternatives on problems that include learning nonlinear stochastic dynamics from video, with or without background distractors.
Our results position the \model as a useful foundation model for a variety of downstream applications.

\end{abstract}
\section{Introduction}

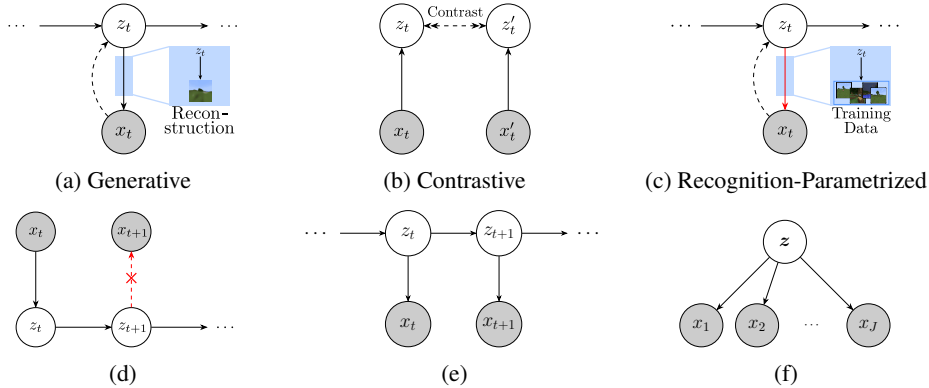
\begin{figure}[t]
    \centering
    \renewcommand{\thesubfigure}{\alph{subfigure}}
    \setlength{\tabcolsep}{2pt}
    \begin{tabular}{ccc}
        \subcaptionbox{Generative\label{fig:generative-diagram}}[0.3\linewidth][c]{%
            \raisebox{-\height}{\includegraphics[height=2cm]{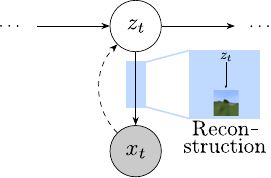}}
        } &
        \subcaptionbox{Contrastive\label{fig:contrastive-diagram}}[0.3\linewidth][c]{%
            \raisebox{-\height}{\includegraphics[height=2cm]{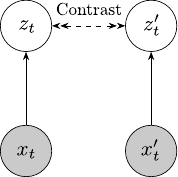}}
        } &
        \subcaptionbox{Recognition-Parametrized\label{fig:recognition-parametrized-diagram}}[0.3\linewidth][c]{%
            \raisebox{-\height}{\includegraphics[height=2cm]{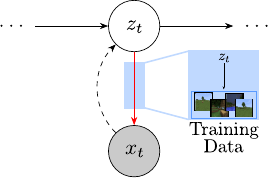}}
        } \\[4ex]
        \subcaptionbox{\label{fig:rollouts-diagram}}[0.3\linewidth][c]{%
            \raisebox{-\height}{\includegraphics[height=1.8cm]{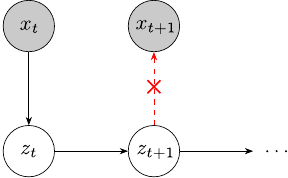}}
        } &
        \subcaptionbox[t]{\label{fig:lds}}[0.3\linewidth][c]{%
            \raisebox{-\height}{\resizebox{0.3\textwidth}{!}{\begin{tikzpicture}[node distance=1.4cm,>=Stealth, every node/.style={minimum size=40pt}]
    \node[draw, circle] (zt) {\LARGE $z_t$};
    \node[draw=none, left=of zt] (ldots) {\LARGE $\cdots$};
    \node[draw, circle, below=of zt, fill=gray!40] (xt) {\LARGE $x_t$};
    \node[draw, circle, right=of zt] (zt2) {\LARGE $z_{t+1}$};
    \node[draw, circle, below=of zt2, fill=gray!40] (xt2) {\LARGE $x_{t+1}$};
    \node[draw=none, right=of zt2] (rdots) {\LARGE $\cdots$};
    
    \draw[->] (ldots) to (zt);
    \draw[->] (zt) to (xt);
    \draw[->] (zt) to (zt2);
    \draw[->] (zt2) to (xt2);
    \draw[->] (zt2) to (rdots);
\end{tikzpicture}}}
        } &
        \subcaptionbox[t]{\label{fig:rpm}}[0.3\linewidth][c]{%
            \raisebox{-\height}{\resizebox{0.2\textwidth}{!}{\begin{tikzpicture}[node distance=1.7cm,>=Stealth,every node/.style={minimum size=33pt}]
    \node[draw, circle] (z) at (-0.25,2.25) {\Large $\bm{z}$};
    
    \node[draw, circle, fill=gray!40] (x1) at (-2.5,0) {\Large $x_1$};
    \node[draw, circle, fill=gray!40] (x2) at (-1,0) {\Large $x_2$};
    \node (dots) at (0.5,0) {$\cdots$};
    \node[draw, circle, fill=gray!40] (xJ) at (2.,0) {\Large $x_J$};

    \draw[->] (z) -- (x1);
    \draw[->] (z) -- (x2);
    \draw[->] (z) -- (xJ);
\end{tikzpicture}}}
        }
    \end{tabular}
    \caption{\textbf{(a)}: Generative methods learn by approximately reconstructing data; \textbf{(b)}: Contrastive methods learn by contrasting representations of data from distinct sequences; \textbf{(c)}: Recognition-parametrized methods learn by implicitly specifying a degenerate decoder that can only generate training data and enforcing conditional independence between observations given latents; \textbf{(d)}: In many applications, future latent variables are predicted, but not future observations; \textbf{(e)}: The SSM graphical model; \textbf{(f)}: Latent variable model with multiple observation factors.}
\end{figure}

Time series are ubiquitous in both machine and natural learning. In applications such as reinforcement learning \citep{DBLP:conf/iclr/0001MCGL21}, robotics \citep{DBLP:conf/icml/ZhangVSA0L19}, navigation \citep{DBLP:conf/icra/CatalJVDS21}, or behavioral neuroscience \citep{friston2021}, agents observe streams of data from which they must infer the state of their environment and, sometimes, predict future states. Observations are often high-dimensional, and agents should be able to infer relevant low-dimensional states in an unsupervised fashion; indeed, this ability is a hallmark of human behavior \citep{lengyel-visual-chunks}. Time series observations are often driven by underlying latent dynamics, formalized by a \textit{state space model} (SSM; \cref{fig:lds}). Most existing SSM learning methods are either \textit{generative} or \textit{contrastive}.%
\blfootnote{
    \textsuperscript{1}Gatsby Unit, UCL
}
\blfootnote{
    \textsuperscript{2}AI Centre, UCL
}
\blfootnote{
    \textsuperscript{$\ast$}Correspondence to: \texttt{s.hromadka@ucl.ac.uk}
}

Generative methods parametrize a joint distribution between latents and observations and maximize the data likelihood (or a lower bound thereof). Crucially, the joint distribution is defined using an explicit likelihood that models how observations are generated from latents.  Generative models in which learning and inference are tractable have limited expressivity; see \cref{subsec:ssms}. Learning more complex models requires approximation.

Deep neural networks have been integrated into state space modeling in recent years, most commonly using the variational autoencoder \citep[VAE;][]{draw,svae}. VAEs are trained using approximate maximum likelihood estimation, enjoying convergence guarantees and principled uncertainty estimation. They work by learning a variational posterior parametrized by a recognition network (or ``encoder''), which is trained concurrently with the generative network (or ``decoder''). Thus, a trained VAE can perform both recognition and generation. However, in many problem domains, such as decision making and control, agents require prediction of future latent states but \textit{not} future observations (\cref{fig:rollouts-diagram}). In such applications, training a generative network is computationally wasteful and can be detrimental to learning. Indeed, VAE-based losses can be interpreted as regularized reconstruction objectives, which encourage latents to encode information about all aspects of the data, including irrelevant distractor features (\cref{fig:generative-diagram}). Another source of error is that generative models that are sufficiently flexible to model real-world data must resort to approximations in inference. These approximations in the variational posterior then bias decoder parameter estimates \citep{turner-sahani-2011}, which in turn directs the encoder towards the wrong generative target \citep{vae-suboptimal}.

Contrastive methods, on the other hand, learn latent representations without modeling how observations are generated. Instead, they train a recognition network to contrast ``positive'' observations from ``negative'' ones (\cref{fig:contrastive-diagram}). Contrastive methods do not learn a full probabilistic state space model, obtaining neither latent dynamics nor posterior beliefs over latent variables---features that are crucial to optimal Bayesian decision making. Recent extensions suggest learning dynamics and uncertainty \citep{contrastive-probabilistic,you2022}, but even these do not restore a fully likelihood-based approach, instead depending on auxiliary losses and regularization.

In this work, we aim to improve state space recognition by combining the best of both worlds: utilizing the likelihood formalism of probabilistic models, but without an explicit generative ``decoder'' (\cref{fig:recognition-parametrized-diagram}). We do this in the framework of the recognition-parametrized model \citep[RPM;][]{rpm}. Applied to an SSM, the RPM focuses on the intuition that observations at different times are conditionally independent given the true latents (\cref{fig:lds}), learning a probabilistic recognition model to infer latents that fit this description. As a result, the learned latents reflect temporally persistent features of the data while systematically ignoring distractors with unreliable statistics across time. We define an RPM with joint Gaussian latent prior that matches the SSM structure, derive the learning algorithm, and demonstrate its efficacy across a range of unsupervised time series datasets. The resulting \model is, to the best of our knowledge, the first fully probabilistic method for modeling latent time series that does not learn an explicit generative model.

\section{Background}

\subsection{Recognition-Parametrized Models}

We first introduce the RPM in its general form, following \citet{rpm}. Consider a latent-variable model where the observation $\bm{x} \coloneqq (x_1,\dots,x_J)$ comprises multiple variables (i.e., $J>1$) that are conditionally independent given the latents (\cref{fig:rpm}):
\begin{equation*}
    x_i \ind x_j \mid \bm{z} \quad \forall i \neq j.
\end{equation*}
The boldface $\bm{z}$ may represent many latent variables with internal conditional independence structure, such as a Markov chain (\cref{sec:rp-lds}).
The graphical model in \cref{fig:rpm} implies a joint distribution on latents $\bm{z}$ and data $\bm{x}$:
\begin{equation*}
    p(\bm{x},\bm{z}) = p(\bm{z}) \prod_{j=1}^J p(x_j|\bm{z}).
\end{equation*}
Equivalently, by Bayes' rule,
\begin{equation*}
    p(\bm{x},\bm{z}) = p(\bm{z}) \prod_{j=1}^J \frac{p(\bm{z}|x_j)p(x_j)}{p(\bm{z})}.
\end{equation*}
Given a dataset $\bm{X} = (x_1^n,\dots,x_J^n)_{n=1}^N$, \citet{rpm} propose to model the true $p(\bm{x},\bm{z})$ with a semi-parametric distribution $\rpmjoint{\bm{x},\bm{z}}$ called the \textit{RPM joint}, given by
\begin{equation} \label{eq:rpm-joint}
    \rpmjoint{\bm{x},\bm{z}} = p_\eta(\bm{z})\prod_{j=1}^J \frac{\rpmfj{\bm{z}|x_j}\rpmpx{x_j}}{\rpmFj{\bm{z}}},
\end{equation}
where $\theta=(\eta,\{\phi_j\}_{j=1}^J)$ and
\begin{itemize}
    \item $p_\eta(\bm{z})$ is a parametrization of the prior;
    \item $\rpmfj{\bm{z}|x_j}$ are parametrized $x_j$-dependent distributions over $\bm{z}$ called \textit{recognition factors};
    \item $\rpmpx{x_j}$ is the empirical data distribution $\rpmpx{x_j} = \frac{1}{N}\sum_{n=1}^N \delta(x_j=x_j^n)$;
    \item $\rpmFj{\bm{z}}$ ensures that the RPM joint is normalized:
    \begin{equation}
        \rpmFj{\bm{z}} = \int \rpmfj{\bm{z}|x_j}\rpmpx{x_j} \diff x_j 
        = \frac{1}{N} \sum_{n=1}^N \rpmfj{\bm{z}|x_j^n}.\label{eqn:mixture}
    \end{equation}
\end{itemize}
The superscript $\bm{X}$ in the RPM joint emphasizes that it is itself a function of the training data through its dependence on the empirical distributions $\rpmpx{x_j}$.

\cref{eq:rpm-joint} defines a joint distribution over observed and latent variables that does not include a parametric component modeling a distribution on observations. Hence, the RPM circumvents learning an explicit generative model, allowing it to focus model capacity on learning good recognition factors. In fact, the generative distribution implied by the RPM joint in \cref{eq:rpm-joint} is
\begin{equation*}
    \rpmjoint{\bm{x}|\bm{z}} \propto \prod_{j=1}^J \rpmfj{\bm{z}|x_j} \rpmpx{x_j},
\end{equation*}
so in a formal sense the RPM can only ever generate values of $x_j$ that appear in the training set. However, as recognition is parametric, the RPM can nonetheless be used to construct posterior beliefs over latents given previously unseen observations.

Notably, the RPM joint is a normalized distribution and can be fit by maximum likelihood using the Expectation Maximization (EM) algorithm and related methods \citep{em-algorithm}. \citet{rpm} implement RPMs in a variety of settings and construct approximations when necessary. We apply some of their methods in \cref{sec:rp-lds}.

\subsection{State Space Models} \label{subsec:ssms}
A probabilistic SSM\footnote{We use the term ``SSM'' to refer a generic probabilistic model with graphical structure as in \cref{fig:lds}, not the deterministic methods in recent literature on structured state space models, e.g. Mamba \citep{mamba}.}
is a latent variable model with Markovian latents $\bm{z} \coloneqq (z_1,\dots,z_T)$, $z_t \in \reals^{\setD_\setZ}$, and noisy, partially informative observations $\bm{x} \coloneqq (x_1,\dots,x_T)$, $x_t \in \reals^{\setD_\setX}$.

The graphical model shown in \cref{fig:lds} provides the joint factorization over both $\bm{x}$ and $\bm{z}$:
\begin{equation} \label{eq:lds-joint}
    p(\bm{x},\bm{z}) = p(z_1) \prod_{t=1}^{T-1} p(z_{t+1}|z_t) \prod_{t=1}^T p(x_t|z_t).
\end{equation}
The standard approach to fitting an SSM to data is to parametrize the model with some parameters $\theta$, define a \textit{variational posterior} $q(\bm{z}) \approx p_\theta(\bm{z}|\bm{x})$, and maximize the \textit{free energy}
\begin{equation*}
    \setF(q,\theta) = \ave{\log p_\theta(\bm{x},\bm{z})}{q(\bm{z})} + \setH(q),
\end{equation*}
where $\ave{\,\cdot\,}{q(\bm{z})}$ denotes expectation with respect to $q$ and $\setH(q)=-\ave{\log q(\bm{z})}{q(\bm{z})}$ is the entropy of $q$. The free energy is a lower bound to the data log-likelihood, $\log p_\theta(\bm{x})$, and can be maximized via the EM algorithm, consisting of alternating updates to $q$ (``E-step'') and $\theta$ (``M-step'') \citep{em-algorithm}. Although the EM algorithm can be performed exactly for discrete latent variables \citep{hmm}---albeit with exponential computational complexity in latent dimensionality---in this work we focus on continuous latent variables. In this setting, there are few models for which EM can be performed exactly.

\subsubsection{Linear-Gaussian SSM}

In a \textit{Gaussian SSM} (GSSM), the components of the prior distribution over the latents take the following jointly Gaussian form:
\begin{align*}
    p(z_1) = \normal{m_1, Q_1}, \qquad
    p(z_t|z_{t-1}) = \normal{Az_{t-1}+b,Q}.
\end{align*}
When paired with linear-Gaussian emissions $p(x_t|z_t) = \normal{Cz_t + d, R}$, the model is called a \textit{linear-Gaussian SSM}. The E-step is solved exactly by Kalman smoothing \citep{kalman-filter} and the M-step can be performed in closed form \citep{ghahramani1996parameter}. 
In general, the parameters $(m_1, Q_1, A, b, Q, C, d, R)$ can vary over time, but we assume them to be time-invariant for notational simplicity. We call $p(z_1)$ the \textit{initial distribution}, $A$ the \textit{transition matrix}, and $b$ the \textit{bias}. Although often chosen for tractability, the combination of linear dynamics and emission distributions is restrictive for many real-world applications. Therefore, in this work, we instead consider extensions to nonlinear-Gaussian emissions: $p(x_t|z_t) = \normal{f(z_t), g(z_t)}$.

\subsubsection{Stability of Gaussian SSMs}

We briefly review the theory of these models, inspired by \citet{stable-lds}. A GSSM is \textit{stable} if $\rho(A) < 1$, where $\rho(\cdot)$ denotes the largest eigenvalue magnitude. A stable system has a well-defined \textit{stationary distribution} $\lim_{t\rightarrow\infty}p(z_t)$ \citep{stable-lds}. The following lemma shows that, in consequence of the non-identifiability of the GSSM parameters, no generality is lost by assuming that a latent process underlying a stable GSSM has zero transition bias and stationary distribution $\normal{0,I}$.
\begin{restatable}[]{lemma}{stat}
    \label{lem:stationary-dist}
    Let the GSSM on observations $\bm{x} = (x_1\dots,x_T)$ and latents $\bm{z} = (z_1,\dots,z_T)$, with parameters $\Theta = (m_1, Q_1, A, b, Q, f, g)$ be stable.
    Then there exists another stable GSSM with parameters $\tilde\Theta = (\tilde m_1, \tilde Q_1, \tilde A, \tilde b, \tilde Q, \tilde f, \tilde g)$ such that
    \begin{equation*}
        p(\bm{x}|\Theta) = p(\bm{x}|\tilde{\Theta})\,,
    \end{equation*}
    the stationary distribution of the latent variables is $\normal{0,I}$, and $\tilde b = 0$.
\end{restatable}
We provide a proof of \cref{lem:stationary-dist} and an explicit form of $\tilde{\Theta}$ in \cref{app:lemma-proof}.

\section{\model} \label{sec:rp-lds}

In this section we apply the RPM framework to the GSSM and develop the necessary approximations to perform learning and inference.

The RPM joint for the SSM graphical model is
\begin{equation*}
    \rpmjoint{\bm{x},\bm{z}} = p_\eta(z_1) \prod_{t=1}^{T-1}p_\eta(z_{t+1}|z_t) \prod_{t=1}^T \frac{\rpmf{\bm{z}|x_t}\rpmpx{x_t}}{\rpmF{\bm{z}}}.
\end{equation*}
Inspired by the factorization of emissions over time in the SSM joint in \cref{eq:lds-joint}, we define the recognition factors as distributions only on $z_t$:
\begin{equation} \label{eq:rp-lds-joint}
    \rpmjoint{\bm{x},\bm{z}} = p_\eta(z_1) \prod_{t=1}^{T-1}p_\eta(z_{t+1}|z_t) \prod_{t=1}^T \frac{\rpmf{z_t|x_t}\rpmpx{x_t}}{\rpmF{z_t}},
\end{equation}
and refer to this form as a recognition-parametrized SSM. The free energy for $N$ time series observations $\bm{X} = (x_1^n,\dots,x_T^n)_{n=1}^N$ is
\begin{align}
    \setF\left(\left\{q\nn\right\}_{n=1}^N, \theta\right) 
      &= \sum_{n=1}^N \Bigg[ 
         \sum_{t=1}^T \Bigg(
           \ave{\log \frac{\rpmf{z_t|x_t\nn}}{\rpmF{z_t}}}{q\nn(z_t)} 
           \!\!\!\!\!\!+ \log \rpmpx{x_t\nn} 
         \Bigg)
    - \kl{q\nn(\bm{z})}{p_\eta(\bm{z})} \Bigg].
    \label{eq:rpm-free-energy}
\end{align}
The explicit $\log p_0$ terms do not depend on $\theta$ or $q\nn$ and so do not affect the optimization. Therefore we omit them below.
A key benefit of recognition parametrization is that model posteriors may be defined to lie within a tractable family even when the link to observations is arbitrarily nonlinear. To keep $q\nn(\bm{z})$ (and $p_\eta(\bm{z})$) tractable, we focus on the GSSM prior.  Inspired by \cref{lem:stationary-dist}, we parametrize the latent process as linear-Gaussian with no bias and a stationary distribution of $\normal{0,I}$. The recognition factors then provide an arbitrary nonlinear link between observations and latents:
\begin{align*}
    p(z_1) = \normal{0,I}, \quad\,\,\,
    p_\eta(z_t|z_{t-1}) = \normal{Az_{t-1},I-AA^\top}, \quad\,\,\,
    \rpmf{z_t|x_t} = \normal{\mu_\phi(x_t), \Sigma_\phi(x_t)}.
\end{align*}
By setting the transition covariance to $I-AA^\top$ we enforce that the GSSM remains in its stationary distribution at every time step. This is a modeling choice and is not necessary to our method. One limitation of the GSSM prior is that the linear dynamics it encodes (before noise) can only have a single isolated fixed point or hyperplane of fixed points, limiting expressivity of latent dynamics. The \model can compensate for this in two ways: by having neural network-parametrized recognition, and by having favorable scaling with latent dimension (due to the efficiency of GSSM inference). Nonlinear dynamics can be approximated by higher-dimensional linear systems, as motivated by Koopman Operator Theory \citep{koopman}.  By combining nonlinear recognition with potentially higher latent dimensionality, the \model can tractably model a wide range of systems, as shown in \cref{sec:experiments}.

Note that $\rpmF{z_t}$ is a mixture of Gaussians (\cref{eqn:mixture}) and so $\ave{\log \rpmF{z_t}}{q^n(z_t)}$ is intractable. Rather than resorting to sample-based Monte Carlo approximations, we resolve this term by adopting the \textit{interior variational bound} \citep{rpm}. Introducing parametrized auxiliary factors $\rpmaux{tn}{\omega}(z_t)$, not necessarily normalized, and applying Jensen's inequality, we get
\begin{align}
    -\ave{\log \rpmF{z_t}}{q^n(z_t)} &= -\ave{\log \frac{\rpmF{z_t}\rpmaux{tn}{\omega}(z_t)q^n(z_t)}{\rpmaux{tn}{\omega}(z_t)q^n(z_t)}}{q^n(z_t)} \nonumber\\
    &\geq \ave{\log \frac{\rpmaux{tn}{\omega}(z_t)}{q^n(z_t)}}{q^n(z_t)} - \log \int \rpmF{z_t}\rpmaux{tn}{\omega}(z_t) \diff z_t. \label{eq:int-var-bound}
\end{align}
If we choose $\rpmaux{tn}{\omega}(z_t)$ to be proportional to a Gaussian, the right-hand side can be computed in closed form. This yields a lower bound to $\setF$, which we call the \textit{auxiliary free energy}, denoted $\setG(\{q^n\}_{n=1}^N, \{\rpmaux{tn}{\omega}\}_{n=1,t=1}^{N,T},\theta)$. The auxiliary free energy can be optimized by alternating maximization of the $q^n$, $\omega$, and $\theta$. However, in practice we found it sufficient to update the $\rpmaux{tn}{\omega}$ approximately, as follows. First, we note that the bound in \cref{eq:int-var-bound} is tight when $\rpmaux{tn}{\omega}(z_t) \propto q^n(z_t) / \rpmF{z_t}$. Moreover, as discussed by \citet{rpm}, $\rpmF{z_t} \rightarrow p_\eta(z_t)$ in the large-data in-model limit. Hence, to support this convergence, we assume $\rpmF{z_t} \approx p_\eta(z_t)$ and set $\rpmaux{tn}{}(z_t) = q^n(z_t) / p_\eta(z_t)$, dropping the $\omega$ subscript as there are no additional parameters to optimize. The auxiliary free energy thus becomes
\begin{align*}
    \setG &\overset{+c}{=} \sum_{n=1}^N \Bigg[ \sum_{t=1}^T \left( \ave{\log \frac{\rpmf{z_t|x_t^n}{\rpmaux{tn}{\omega}(z_t)}}{q^n(z_t)}}{q^n(z_t)} - \log\Gamma^{tn}_{\phi_t} \right) - \kl{q^n(\bm{z})}{p_\eta(\bm{z})}\Bigg],
\end{align*}
where the terms $\Gamma^{tn}_{\omega,\phi_t}$ can be computed in closed form. A full derivation of $\mathcal{G}$ and a discussion of the variational gap introduced by the interior variational bound are provided in \cref{app:loss-derivation}. Since $p(z_t|x_t) \propto p(z_t) p(x_t|z_t)$ (by Bayes' rule), we further choose to parametrize $\rpmf{z_t|x_t} \propto p_\eta(z_t) \rpmfdelta(z_t|x_t)$, where the $\rpmfdelta{}$ are learned time-invariant distributions. The auxiliary free energy can then be rewritten as
\begin{align*}
    \setG &\overset{+c}{=} \sum_{n=1}^N \Bigg[ \sum_{t=1}^T \left( \ave{\log \rpmfdelta(z_t|x_t^n)}{q^n(z_t)} - \log\tilde{\Gamma}^{tn}_{\phi_t} \right) - \kl{q^n(z_t)}{p_\eta(z_t)} \Bigg].
\end{align*}
The \model is trained via standard EM on $\setG$.
Up to an additive constant independent of the $q^n$, $\mathcal{G}$ has the same form as the free energy of a linear-Gaussian system with emission factor $p(x_t\nn|z_t)$ replaced by $\rpmfdelta(z_t|x_t\nn)$. As we define the $\rpmfdelta{}$ distributions on $z_t$ to be Gaussian, $q\nn(z_t)$ can be computed \textit{exactly} via standard Kalman smoothing. Furthermore, as all terms in $\setG$ can be computed exactly, the M-step is achieved by gradient ascent on $\setG$ with respect to $\theta$.

The \model supports exact inference and avoids Monte Carlo approximations. These unique properties provide graceful scaling with latent dimensionality, as we show in \cref{subsec:linear-task}.

\section{Related Work} \label{subsec:related-work}
Early extensions of Kalman smoothing to nonlinear transition and emissions distributions include the Extended Kalman filter \citep[EKF;][]{extended-kalman-filter}, which linearizes around posterior means, and the Unscented Kalman filter \citep{unscented-kalman-filter}, which builds on the EKF by using a sampling-based approach that allows approximation up to third order.

One of the first successful VAE-based time series models was the Deep Kalman Filter \citep[DKF;][]{krishnan2015deepkalmanfilters}. The DKF parametrizes the variational posterior $q$ with a recurrent neural network (RNN) and uses reparametrization to take gradients through a sample-estimated free energy. The Kalman VAE \citep[KVAE;][]{DBLP:conf/nips/FraccaroKPW17} and Recurrent Kalman Network \citep[RKN;][]{becker2019recurrent}, on the other hand, combine \textit{locally} linear latent dynamics with VAE-based generative networks. The KVAE utilizes the local linearity to enable Kalman smoothing-based posterior inference, whereas the RKN combines local linearity with factorized state beliefs to learn high-dimensional latent state representations efficiently.

Perhaps the generative method closest to the \model is the structured VAE \citep[SVAE;][]{svae,zhao2023revisiting}, which learns a recognition network that outputs conjugate potentials on latents. When paired with linear-Gaussian transitions, the potentials allow for exact inference via Kalman smoothing. However, unlike the \model, the SVAE also learns an explicit generative model using reparametrization.

Contrastive Predictive Coding (CPC) \citep{infonce} exploits the contrastive InfoNCE loss to identify a latent representation that maximizes internal predictability. Here, a recognition network summarizing past data in a sequence aims to discriminate future latents from the same sequence (positive) from draws taken from other sequences (negative). Recent extensions of contrastive learning provide ways to learn explicit latent dynamics by adding an auxiliary loss \citep{you2022} or by integrating parametrized dynamics directly into the model's similarity function \citep{laiz2025selfsupervised}.

\section{Experiments} \label{sec:experiments}

We compare the \model to various generative and contrastive approaches for learning latent time series from high-dimensional observations.
We show that the \model is able to outperform baselines at recovering accurate latent representations across all tasks, and even more so when background distractions are present in the observed data.
These results point to the benefits of the tractable prior combined with nonlinear recognition that is not biased by the need to track an explicit generative model.
We also train an \textit{auxiliary} generative model on the \model latents, which we show is able to ``filter out'' the distractions, which generative or contrastive approaches are unable to do.
Additionally, we demonstrate the \model's ability to learn accurate predictive dynamics without the need for auxiliary objectives or masking procedures employed by other methods.

\subsection{Baseline Methods, Hyperparameters, and Estimation}

We compare results from the \model against a variety of generative and contrastive baseline methods: SVAE, KVAE, DKF with four different parametric posterior families, and CPC. Full details are presented in Appendices \ref{app:implementation} and \ref{app:experiments}.

We ran a hyperparameter search for each model on each task, as detailed in \cref{app:hyperparams}. We measured the accuracy of latent recovery by the linear regression $R^2$ between each model's posterior means and the ground truth dynamical variables of each system---arguably the simplest possible downstream task. Results using nonlinear kernel ridge regression, which are qualitatively similar, are provided in \cref{app:further-results}. We apply self-supervised masking to all baselines as described by \citet{zhao2023revisiting}, which was shown to improve predictive performance by encouraging dynamics learning. We found that the \model does not need such schemes to learn the latent dynamics well.

\begin{figure}[t]
    \centering
    \includegraphics[width=0.8\textwidth]{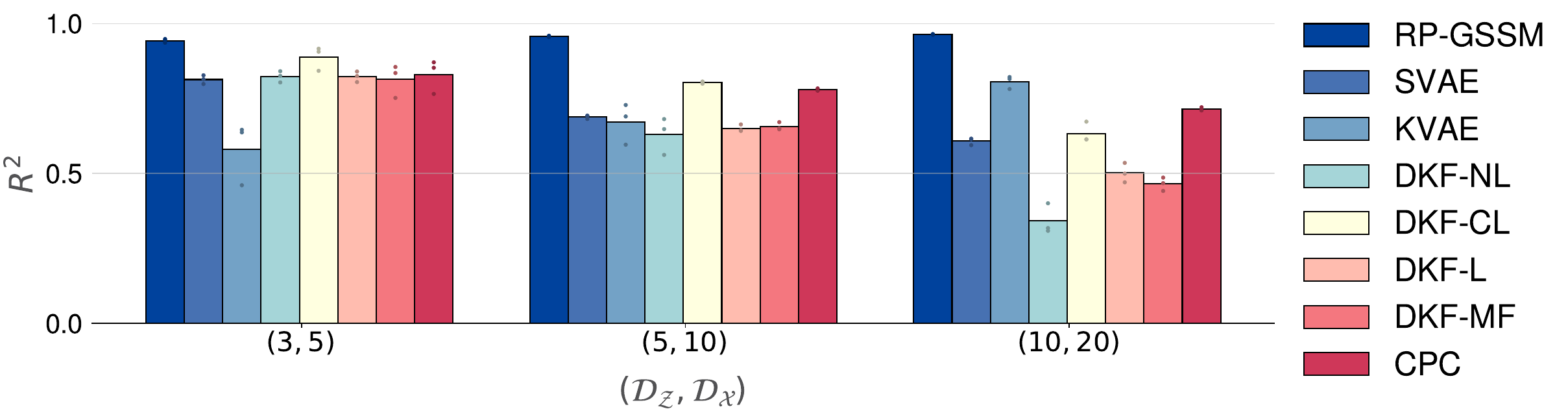}
    \caption{Linear regression $R^2$ score from posterior means to ground truth latent variables for the linear dynamical system, across different latent and observation dimensions. Each dot corresponds to a different seed used to generate the data.}
    \label{fig:linear}
\end{figure}

\subsection{Linear Dynamical System with Linear Emissions} \label{subsec:linear-task}
We begin with data sampled from a linear GSSM (i.e., linear-Gaussian dynamics and emissions):
\begin{gather*}
    z_1 \sim \setN(0, I), \qquad  z_t \sim \setN(Bz_{t-1}, I-BB^\top), \qquad
    x_t \sim \setN(Cz_t+d, R)\,.
\end{gather*}
$B$ was a scaled rotation matrix, applying a rotation of $\pi/5$ around a randomly chosen axis, with operator norm $0.95$; $C$ and $d$ had components sampled i.i.d from $\normal{0,1}$, and $R=0.3I$.  All recognition and generative networks were taken to be linear.  Even in this simple setting, the \model outperforms baselines, particularly for larger dimensionalities (\cref{fig:linear}).

\begin{figure*}[t]
    \centering
    \includegraphics[width=\textwidth]{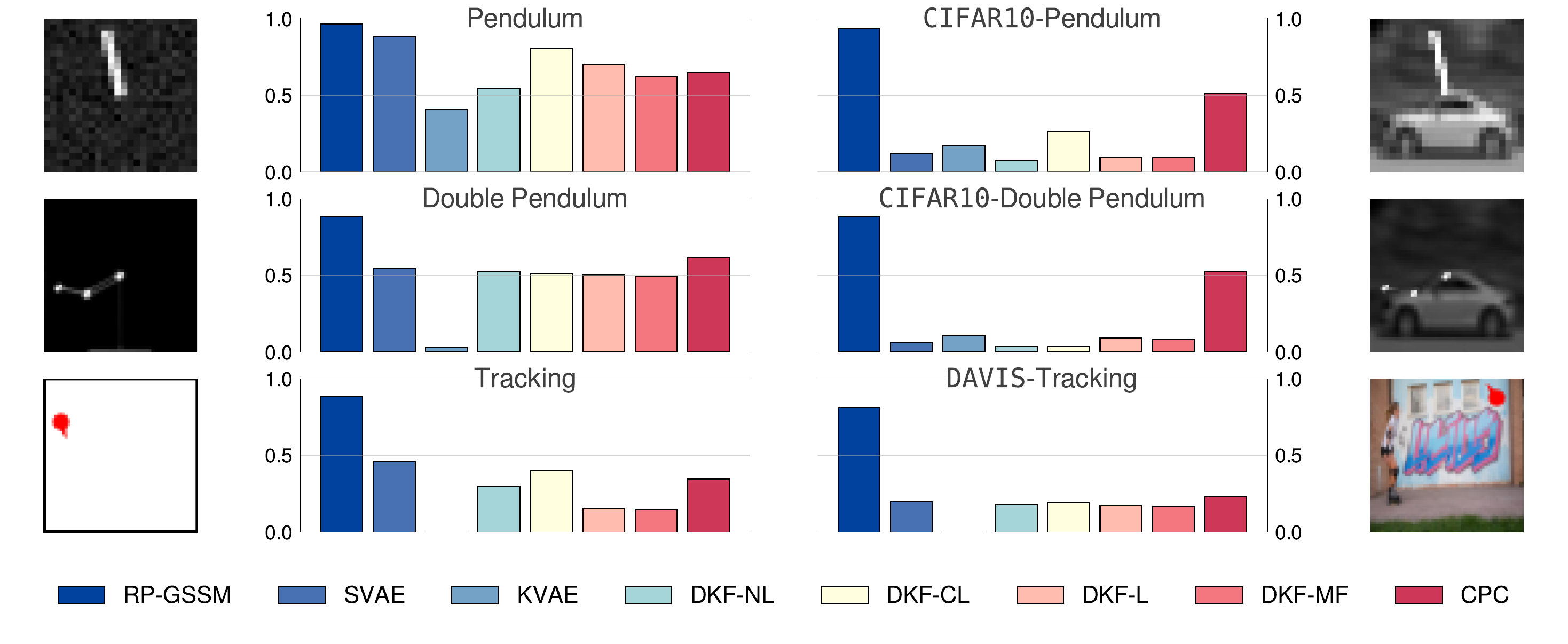}
    \caption{The mean linear regression $R^2$ score between inferred posterior means and true latent state variables. We find the \model learns more informative latent representations than baselines in both the absence (left) and the presence (right) of background distractors.}
    \label{fig:r2s}
\end{figure*}

\subsection{Pendulum and Double Pendulum}
Although consideration of in-model linear dynamics is instructive, a more realistic test of the \model is whether it can model systems with ground truth nonlinear dynamics and nonlinear observation emissions. We applied all models to datasets of noisy synthetic videos of a single pendulum \citep{becker2019recurrent} and real videos of a double pendulum \citep{asseman2018learning}, both evolving freely under gravity from random initial conditions (see \cref{app:experiments} for details).
All recognition and generative networks are three-layer convolutional neural networks.

The true underlying dynamics of a pendulum are governed by angle and angular velocity, which, coupled with fixed linear acceleration due to gravity, form a nonlinear dynamical system. The dynamics of a double pendulum are higher-dimensional, nonlinear, and chaotic. The linear regression targets we use for the pendulum are sine of angle (to avoid discontinuities) and angular velocity. For the double pendulum we use the sine of the angle of the primary pendulum relative to the origin and the sine of the angle of the secondary pendulum relative to the primary pendulum. The mean linear $R^2$ scores to all targets are shown in the top two rows (left) of \cref{fig:r2s}. Despite the ground truth dynamics being nonlinear, all methods perform reasonably well at decoding the targets, with the \model being most consistent.

The ability of the \model to perform exact inference without having to approximate a complex generative network allows for increased data efficiency. \cref{fig:pendulum-small} in \cref{app:further-results} shows the results of all methods trained on only the first 64 sequences of the single-pendulum dataset. The \model is the only model able to linearly decode angular velocity after learning on such a small dataset.

\subsubsection{\texttt{CIFAR10} Background Distractions}
To demonstrate the performance of our model in the presence of background distractions, we apply the \model to duplicate datasets of the pendulum and double-pendulum problems, but with randomly selected images from the \texttt{CIFAR10} dataset \citep{krizhevsky2009learning} in the background at each time step. We refer to these datasets as \texttt{CIFAR10}-Pendulum and \texttt{CIFAR10}-Double Pendulum.

We regress to the same targets as in the non-distractor setting. Results are shown in the top two rows (right) of \cref{fig:r2s}, and they clearly show that only the \model is able to reliably decode the underlying dynamical system in the presence of distractors. While the generative baselines use model capacity on learning features of the observation backgrounds, impeding their latent state inference, the \model ignores this and focuses on learning optimal recognition and latent dynamics models.

\begin{figure}[t]
    \centering
    \setlength{\tabcolsep}{2pt}
    \begin{tabular}{ccc}
        \multicolumn{2}{c}{%
            \multirow{2}{*}[2.7cm]{%
                \subcaptionbox*{}{%
                    \begin{overpic}[width=0.5\textwidth]{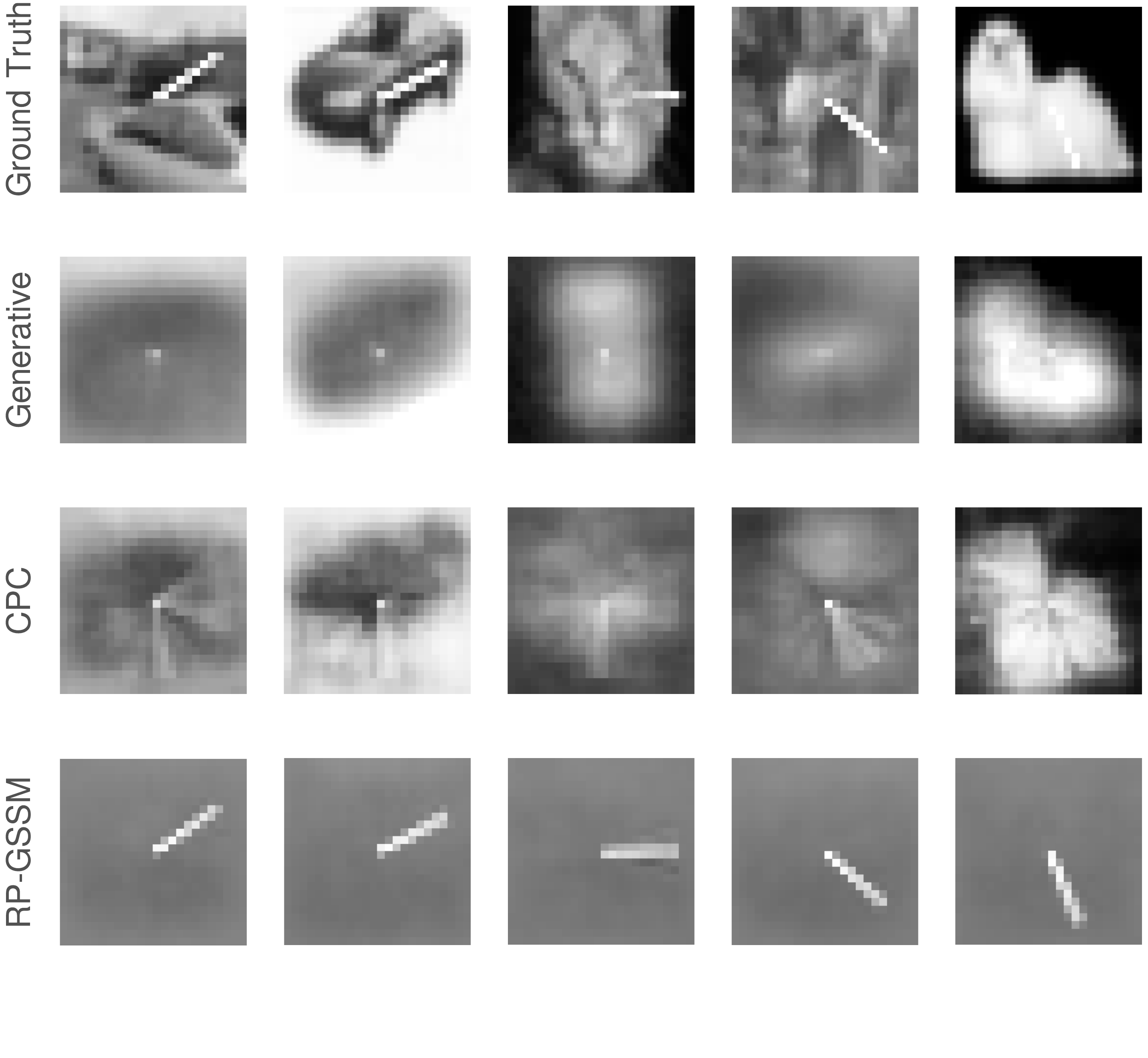}\put(-5,95){(a)}
                    \end{overpic}
                }
            }
        } &
        \subcaptionbox*{}{%
        \begin{overpic}[width=0.4\textwidth]{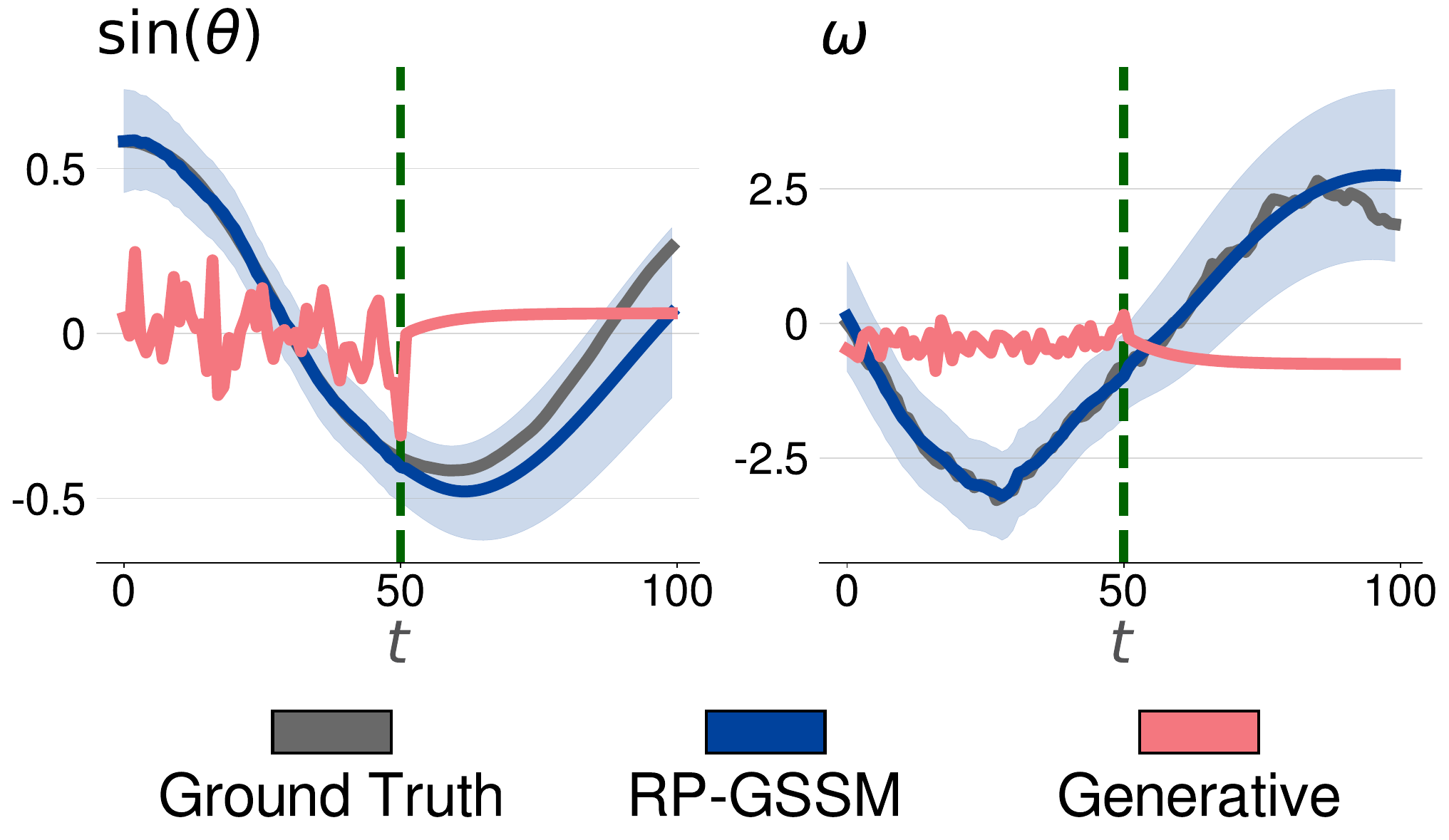}\put(-3,57){(b)}
        \end{overpic}
        } \\
        & &
        \subcaptionbox*{}{%
            \begin{overpic}[width=0.4\textwidth,trim={4cm 4cm 2.8cm 4cm},clip]{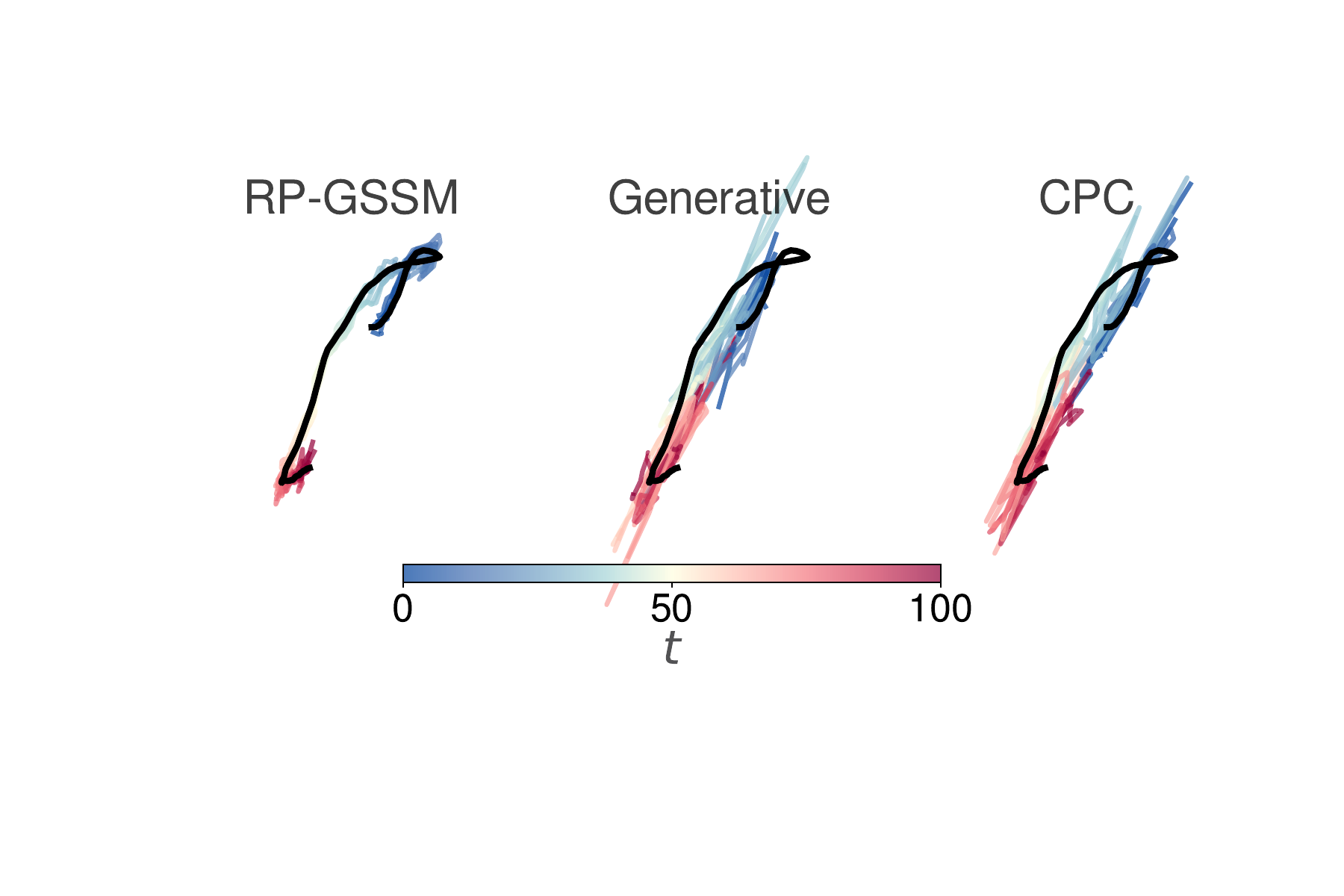}\put(-3,55){(c)}
            \end{overpic}
        }
    \end{tabular}
    \vspace{-0.6cm}
    \caption{\textbf{(a)}: Reconstructions on \texttt{CIFAR10}-Pendulum for the best generative baseline and auxiliary generative models trained using the best CPC and \model models. The \model is the only model that allows accurate reconstruction of the pendulum arm. \textbf{(b)}: When given 50 \texttt{CIFAR10}-Pendulum frames as context, the \model is able to infer the true states during the context and predict the next 50 states. The best generative baseline is unable to do either. The dashed line represents when the context ends. \textbf{(c)}: The \model performs more accurate tracking in the \texttt{DAVIS}-Tracking dataset than the best generative and CPC baselines. The black line is a sample ground truth trajectory and the colored lines are inferred trajectories across five different background videos.}
    \label{fig:applications}
\end{figure}

To illustrate the \model's ability to extract temporally consistent signal while ignoring distractors, we train an auxiliary generative model using the learned posteriors from the best \model and CPC runs on \texttt{CIFAR10}-Pendulum. We compare reconstructions from these models with those of the best performing generative baseline in \subfigref{fig:applications}{a}. Notably, only the RP-GSSM can accurately reconstruct the pendulum while ignoring irrelevant background distractions. Further details are presented in \cref{app:implementation}. We also show that the \model can be used to predict future dynamics while ignoring distractors. A sample predicted trajectory of angles and angular velocities on the \texttt{CIFAR10}-Pendulum dataset is shown in \subfigref{fig:applications}{b}.

\subsection{Object State Tracking}

Finally, we apply the \model to object state tracking from video. This task is of particular interest when it comes to real-world applications, as state tracking from visual observations is a common problem for both biological agents and in robotics.

The dataset comprises a set of synthetic video sequences simulating a rat running around a box \citep{George2024} (see \cref{app:experiments} for details). The true dynamics are governed by an Ornstein-Ulhenbeck process on the velocity vector. Position changes linearly with velocity across time steps. Although these dynamics are linear, the learned models have to deal with discontinuities in the form of barriers at the walls of the ``box'', as well as having to learn from high-dimensional observations.

\subsubsection{Tracking with Video Backgrounds}
Similar to the \texttt{CIFAR10}-Pendulum and -Double Pendulum datasets above, we apply the \model to a tracking dataset with background distractors. However, this time we apply distractions that are correlated across time, in the form of real-world videos. We stitch together randomly selected 2s video clips from the \texttt{DAVIS} dataset \citep{perazzi2016benchmark, Pont-Tuset_arXiv_2017} and superimpose the simulated object moving on top. We refer to this dataset as \texttt{DAVIS}-Tracking.

In all tracking experiments, linear regression targets were the position, velocity, and head direction (low-pass filtered velocity, red arrow in \cref{fig:r2s}, bottom right) of the ``rat'' being tracked, all of which are two-dimensional. The mean $R^2$ values across all targets are shown in the bottom row of \cref{fig:r2s}. The \model is the only method able to consistently track the ``rat''. In \subfigref{fig:applications}{c} we visualize the inferred latent states for a single trajectory with 5 different background videos. The results demonstrate both the ability of the \model to scale to high-dimensional data due to its lack of parametric generative model, as well as its robustness to changes in background distractor---something the generative and contrastive methods do not share.

\section{Conclusion}

Traditional probabilistic methods rely on concurrently training parametric recognition and generative models. In contrast, the \model specifies the ``generative'' model implicitly in terms of the recognition process. As a result, learning is not driven by a reconstruction objective, but rather by aligning recognition and inference with the specified prior structure, i.e. by finding latent representations that capture dependencies between different observational factors. We rely on the fact that time series data naturally provides such factors in the form of observations at different time steps, and posit that ``good'' latent representations are those that account for the temporal structures and correlations in the observed time series. Indeed, our experiments show that this learning objective is useful for producing latent representations that more reliably filter out distractors in the observations, even if these distractors have their own internal structure (e.g. natural images and videos).

Our approach heavily relies on linear dynamical systems as the prior class for the latent time series, which makes efficient inference in the latent space possible via Kalman smoothing. Nevertheless, its applications are not limited to systems that are strictly linear. There are two sources for such nonlinearity. The first is the link between latents and observations, which can be highly nonlinear even if the true latents evolve according to linear dynamics. The second is the underlying dynamics themselves, which can be nonlinear. In both cases, the recognition network can be thought of as finding an ``embedding'' of the observations such that the dynamics of these embeddings can be described by a linear system. These can be of higher dimensionality than the true latents, but are still compressed relative to the full high-dimensional observations.

The general idea behind the \model can be extended to more complicated graphical models. One future direction of work is to extend the \model to switching dynamical systems \citep{DBLP:conf/icml/DongS0B20}. Another is to integrate control inputs into the latent dynamical system, which can be applied to simultaneous localization and mapping \citep{DBLP:conf/icra/CatalJVDS21}, optimal control \citep{DBLP:conf/corl/WatsonA019}, and model-based reinforcement learning \citep{DBLP:journals/corr/abs-1805-00909}. Finally, additional data modalities such as sound, text, or reward can be incorporated into the graphical model. The resulting increase in the number of conditional independence relationships in the graph would strengthen the \model learning signal, improving its ability to disregard irrelevant features.

\newpage
\bibliography{ref}
\bibliographystyle{apalike}

\newpage
\appendix

\section{Proof of \cref{lem:stationary-dist}} \label{app:lemma-proof}

\stat*

\textit{Proof. }
The GSSM with parameters
\begin{equation*}
    \Theta = (m_1, Q_1, A, b, Q, f, g)
\end{equation*}
defines a joint distribution on  $(z_1,\dots,z_T)$ and $(x_1,\dots,x_T)$ according to 
\begin{align*}
    z_1 &\sim \normal{m_1,Q_1},\\
    z_t | z_{t-1}&\sim\normal{Az_{t-1} + b, Q} && \text{for } t=2\dots T, \text{and} \\
    x_t | z_t &\sim\normal{f(z_t),g(z_t)}, && \text{for } t=1\dots T.
\end{align*}
Let $G$ be invertible, and consider the variables $u_t \coloneqq Gz_t + c$ for all $t$.  Using the properties of Gaussian distributions under affine transformations of variables we have
\begin{align*}
    u_1 &\sim \normal{G m_1 + c, GQ_1G\tr}
\intertext{and}
    u_t | z_{t-1} &\sim\normal{GAz_{t-1} + c + Gb, GQG\tr}
\intertext{or, noting that $z_t = G\inv(u_t - c)$,}
    u_t | u_{t-1} &\sim\normal{GAG\inv u_{t-1} + (I-GAG\inv) c + Gb, GQG\tr} && \text{for } t=2\dots T.
\intertext{Furthermore,}
    x_t | u_t &\sim \normal{f\left(G^{-1}(u_t-c)\right), g(G^{-1}(u_t-c)},
\intertext{or}
    x_t | u_t &\sim \normal{(f \circ h)(u_t), (g \circ h)(u_t)},
\end{align*}
where $h : u_t \mapsto G^{-1}(u_t-c)$.

Therefore, the variables $(u_1,\dots,u_T)$ and emissions $(x_1,\dots,x_T)$ are jointly distributed according to a GSSM with parameters
\begin{equation} \label{eq:theta-tilde}
    \tilde{\Theta} \coloneqq \left( Gm_1+c, GQ_1G^\top, GAG^{-1}, \left(I-GAG^{-1}\right)c+Gb, GQG^\top, f \circ h, g \circ h \right)\,.
\end{equation}
The emission variables were left unchanged by the transformation, and so 
\begin{equation*}
    p(\bm{x}|\Theta) = p(\bm{x}|\tilde{\Theta}).
\end{equation*}
The transition matrices on $z_t$ and $u_t$, given by $A$ and $GAG^{-1}$ respectively, are similar and hence have the same eigenvalues. Therefore $\rho(A) = \rho(GAG^{-1})$ and the GSSM given by $\tilde{\Theta}$ for any invertible $G$ is stable.

Next, let the stationary distribution of $z_t$ be $\normal{m_\infty, Q_\infty}$. The stationary distribution $p_\infty$ satisfies
\begin{equation*}
    p_\infty(z_{t+1}) = \int p(z_{t+1}|z_t) p_\infty(z_t) \diff z_t,
\end{equation*}
which yields the \textit{Lyapunov equation}:
\begin{equation}
    Q_\infty = A Q_\infty A^\top + Q,
\end{equation}
as well as an equation for the stationary mean:
\begin{equation} \label{eq:stat-mean}
    m_\infty = Am_\infty + b.
\end{equation}

Denote the stationary distribution of the system with parameters $\tilde{\Theta}$ by $\normal{\tilde{m}_\infty,\tilde{Q}_\infty}$. $\tilde{Q}_\infty$ satisfies
\begin{align*}
    \tilde{Q}_\infty &= G Q_\infty G^\top.
\end{align*}
$Q_\infty$ is symmetric positive definite and can be decomposed as $Q_\infty = USU^\top$ where $U$ is orthogonal and $S$ is diagonal. Setting $G \coloneqq S^{-1/2}U^\top$ gives
\begin{equation*}
    \tilde{Q}_\infty = I.
\end{equation*}
Next, note that in \cref{eq:stat-mean}, if the bias $b$ is $0$, then $Am_\infty = m_\infty$. However, this would imply that $A$ has an eigenvalue of $1$; a contradiction. Therefore, $b=0 \implies m_\infty=0$.

The bias of the system defined by $\tilde{\Theta}$, given in \cref{eq:theta-tilde}, is $\left(I-GAG^{-1}\right)c + Gb$. Setting to $0$ and solving for $c$,
\begin{equation*}
    c = \left(GAG^{-1}-I\right)^{-1}Gb.
\end{equation*}
This equation has a unique solution for $c$ if and only if $\left| GAG^{-1}-I \right| \neq 0$. This is the case if and only if $GAG^{-1}$ does not have an eigenvalue equal to $1$. However, as established previously, $GAG^{-1}$ has eigenvalues equal to those of $A$, which cannot be $1$ by assumption. Therefore, there exists a vector $c$ such that the system defined by $\tilde{\Theta}$ has zero bias, and therefore, as argued above, also has stationary mean $\tilde{m}_\infty$ equal to $0$. This completes the proof. \qed

\section{The \model Loss} \label{app:loss-derivation}

\subsection{Variational Gaps}

The \model free energy, as defined in \cref{eq:rpm-free-energy}, is given by
\begin{align}
    \setF\left(\left\{q\nn\right\}_{n=1}^N, \theta\right) 
      &= \sum_{n=1}^N \Bigg[ 
         \sum_{t=1}^T \Bigg(
           \ave{\log \frac{\rpmf{z_t|x_t\nn}}{\rpmF{z_t}}}{q\nn(z_t)} 
           \!\!\!\!\!\!+ \log \rpmpx{x_t\nn} 
         \Bigg)
    - \kl{q\nn(\bm{z})}{p_\eta(\bm{z})} \Bigg].
    \tag{\ref{eq:rpm-free-energy}}
\end{align}
The free energy is a lower bound to the log-likelihood of the data under the \model joint distribution (\cref{eq:rpm-joint}):
\begin{equation*}
    \setF\left(\left\{q\nn\right\}_{n=1}^N, \theta\right) \leq \sum_{n=1}^N \log \rpmjoint{\bm{x}^n},
\end{equation*}
where $\bm{x}^n = (x_1^n,\dots,x_T^n)$. The difference between the data log-likelihood and the free energy is called a \textit{variational gap} and can be written explicitly:
\begin{equation*}
    \log \rpmjoint{\bm{x}^n} - \setF\left(\left\{q\nn\right\}_{n=1}^N, \theta\right) = \sum_{n=1}^N \kl{q^n(\bm{z})}{\rpmjoint{\bm{z}|\bm{x}^n}}.
\end{equation*}
Using the interior variational bound described in \cref{eq:int-var-bound}, we obtain the \model auxiliary free energy:
\begin{align}
    \setG\left(\{q^n\}_{n=1}^N, \{\rpmaux{tn}{\omega}\}_{n=1,t=1}^{N,T},\theta\right) &= \sum_{n=1}^N \Bigg[ \sum_{t=1}^T \left( \ave{\log \frac{\rpmf{z_t|x_t^n}{\rpmaux{tn}{\omega}(z_t)}}{q^n(z_t)}}{q^n(z_t)} - \log\Gamma^{tn}_{\phi_t} \right) \nonumber\\ &\qquad\qquad- \kl{q^n(\bm{z})}{p_\eta(\bm{z})}\Bigg] - TN\log N \label{eq:aux-free-energy} \\
    &\leq \setF\left(\left\{q\nn\right\}_{n=1}^N, \theta\right), \nonumber
\end{align}
where the terms $\log\Gamma^{tn}_{\phi_t}$ are given explicitly in \cref{eq:int-var-bound}:
\begin{align}
    \Gamma^{tn}_{\phi_t} &= \int \rpmF{z_t}\rpmaux{tn}{\omega}(z_t) \diff z_t \nonumber\\
    &= \frac{1}{N} \sum_{n'=1}^N \int \rpmf{z_t\Big|x_t^{n'}} \rpmaux{tn}{\omega}(z_t) \diff z_t. \label{eq:Gamma-terms}
\end{align}
The gap induced by using Jensen's inequality in \cref{eq:int-var-bound} can also be written explicitly:
\begin{align*}
    \log \rpmjoint{\bm{x}^n} \,-\, &\setG\left(\{q^n\}_{n=1}^N, \{\rpmaux{tn}{\omega}\}_{n=1,t=1}^{N,T},\theta\right) \\ &= \sum_{n=1}^N \kl{q^n(\bm{z})}{\rpmjoint{\bm{z}|\bm{x}^n}} - \sum_{n=1}^N\sum_{t=1}^T \kl{q^n(z_t)}{\frac{\rpmF{z_t}\rpmaux{tn}{\omega}(z_t)}{\Gamma^{tn}_{\phi_t}}}.
\end{align*}
The former variational gap vanishes when $q^n$ approaches the true posterior implied by the \model joint. The latter gap vanishes when $\rpmaux{tn}{\omega}(z_t) \propto q^n(z_t)/\rpmF{z_t}$, as also implied by Jensen's inequality in \cref{sec:rp-lds}.

\subsection{Loss Derivation}

The full \model auxiliary free energy is provided above in \cref{eq:aux-free-energy}. Although in the main text we assume Gaussian distributions throughout, here we allow any exponential family with constant base measure for generality (dropping parameter subscripts for clarity):
\begin{align*}
    f(z_t|x_t^n) &= e^{\eta(x_t^n)^\top t(z_t) - \Phi(\eta(x_t^n))} \\
    q^n(z_t) &= e^{(\eta_{qt}^n)^\top t(z_t) - \Phi(\eta_{qt}^n)} \\
    p(z_t) &= e^{\eta_{0t}^\top t(z_t) - \Phi(\eta_{0t})}.
\end{align*}
I.e., the terms above belong to the same exponential family but with different natural parameters. Let the auxiliary factors $g^{tn}(z_t)$ have the same shape and their own natural parameters $\tilde{\eta}_t^n$:
\begin{equation*}
    g^{tn}(z_t) = e^{(\tilde{\eta}_t^n)^\top t(z_t)}.
\end{equation*}
The auxiliary factors do not have to be normalized, so we model them without log-normalizer ($\Phi$) terms. With this parametrization, the $\Gamma^{tn}$ terms from \cref{eq:Gamma-terms} become
\begin{align}
    \Gamma^{tn} &= \frac{1}{N} \sum_{n'=1}^N \int e^{\left(\eta\left(x_t^{n'}\right) + \tilde{\eta}_t^{n}\right)^\top t(z_t) - \Phi\left(\eta\left(x_t^{n'}\right)\right)} \diff z_t \nonumber\\
    &= \frac{1}{N} \sum_{n'=1}^N e^{\Phi\left( \eta\left(x_t^{n'}\right) + \tilde{\eta}_t^{n} \right) - \Phi\left( \eta\left(x_t^{n'}\right) \right)}. \label{eq:Gamma-terms-expfam}
\end{align}
In \cref{sec:rp-lds} we introduced the parametrizations $g^{tn}(z_t) \propto q^n(z_t)/p(z_t)$ and $f(z_t|x_t^n) \propto p(z_t) f^\Delta(z_t|x_t^n)$. In the interior variational bound we multiply and divide by $g^{tn}$, so these terms can be multiplied by arbitrary constants which cancel out. The terms $f(z_t|x_t^n)$, on the other hand, are normalized to be valid probability distributions. In natural parameter space the equivalent parametrizations are
\begin{align*}
    \tilde{\eta}_t^n &= \eta_{qt}^n - \eta_{0t} \\
    \eta(x_t^n) &= \eta_{0t} + \eta^\Delta(x_t^n),
\end{align*}
where $\eta^\Delta (x_t^n)$ are the natural parameter corresponding to $f^\Delta(z_t|x_t^n)$ (which is a member of the same exponential family as all other distributions above). In addition to supporting the convergence $F(z_t) \rightarrow p(z_t)$ in the large-data in-model limit, this parametrization ensures that the natural parameters in \cref{eq:Gamma-terms-expfam}, i.e., $\eta(x_t^{n'})$ and $\eta(x_t^{n'})+\tilde{\eta}_t^n$, are valid natural parameters for all $n, n', t$. In particular, when all distributions are Gaussian, the parametrization ensures that all covariance matrices are positive definite, avoiding singularities.

The normalizing constant for $f(z_t|x_t^n)$, denoted $Z^{tn}$, is given by
\begin{equation*}
    f(z_t|x_t^n) = \underbrace{e^{- \Phi(\eta_{0t}+\eta^\Delta(x_t^n)) + \Phi(\eta_{0t}) + \Phi(\eta^\Delta(x_t^n))}}_{1/Z^{tn}} p(z_t) f^\Delta (z_t|x_t^n).
\end{equation*}
Putting everything together, the interior variational bound is
\begin{align*}
    \ave{\log \frac{f(z_t|x_t^n)}{F(z_t)}}{q^n(z_t)} &\geq \ave{\log \frac{f(z_t|x_t^n)g^{tn}(z_t)}{q^n(z_t)}}{q^n(z_t)} - \log\Gamma^{tn} \\
    &= \ave{\log f^\Delta(z_t|x_t^n)}{q^n(z_t)} - \log Z^{tn} - \log\Gamma^{tn} \\
    &= \ave{\log f^\Delta(z_t|x_t^n)}{q^n(z_t)} - \log\tilde{\Gamma}^{tn},
\end{align*}
where
\begin{equation*}
    \log\tilde{\Gamma}^{tn} = \Phi(\eta_{0t}+\eta^\Delta(x_t^n)) - \Phi(\eta_{0t}) - \Phi(\eta^\Delta(x_t^n)) - \log N + \sum_{n'=1}^N e^{\Phi(\eta_{qt}^n + \eta^\Delta(x_t^{n'})) - \Phi(\eta_{0t} + \eta^\Delta(x_t^{n'}))}.
\end{equation*}
This yields the final loss presented in \cref{sec:rp-lds}:
\begin{equation*}
    \setG\left(\{q^n\}_{n=1}^N, \theta\right) = \sum_{n=1}^N \left[ \sum_{t=1}^T \left( \ave{\log \rpmfdelta(z_t|x_t^n)}{q^n(z_t)} - \log \tilde{\Gamma}_{\eta,\phi}^{tn} \right) - \kl{q^n(\bm{z})}{p_\eta(\bm{z})} \right],
\end{equation*}
where $\theta = (\eta, \phi)$. The terms $\rpmaux{tn}{\omega}$ are dropped from the arguments of $\setG$ because they are now defined in terms of $\theta$ and the $q^n$.

\section{Implementation Details} \label{app:implementation}
All implementation was done in \texttt{jax} \citep{jax}, using the \texttt{dynamax} package \citep{dynamax} for Kalman smoothing.

\subsection{Baselines}
Here we provide details of the baseline models used in this work. We follow \citet{zhao2023revisiting} in the implementation of the SVAE and DKF baselines, and our code for these was adapted from their code repository. These models have a linear-Gaussian prior chain over the latents and neural-network parametrized Gaussian emissions with constant diagonal covariance:
\begin{align*}
    z_1 &\sim \setN(m_1, Q_1) \\
    z_t|z_{t-1} &\sim \setN(Az_{t-1} + b, Q) \\
    x_t|z_t &\sim \setN(\mu_\theta(z_t), \sigma^2
    I)
\end{align*}
The four DKF baselines differ in their variational posterior families. These are as follows:
\begin{align*}
\intertext{\textbf{DKF-MF:}}
    q_\phi(z_{1:T}) &= \prod_{t=1}^T \setN(z_t|\mu_{\phi, t}(x_{1:T}), \Sigma_{\phi, t}(x_{1:T}))
\intertext{where $\{\mu_{\phi, t}(x_{1:T}), \Sigma_{\phi, t}(x_{1:T})\}_{t=1}^T$ are the outputs of a bi-directional RNN.}
\intertext{\textbf{DKF-L:}}
    q_\phi(z_{1:T}) &= \setN(z_t|\mu_{\phi, 1}(x_{1:T}), \Sigma_{\phi, 1}(x_{1:T})) \prod_{t=2}^T \setN(z_t|A z_{t-1} + \mu_{\phi, t}(x_{1:T}), \Sigma_{\phi, t}(x_{1:T}))
\intertext{
where $\{\mu_{\phi, t}(x_{1:T}), \Sigma_{\phi, t}(x_{1:T})\}_{t=1}^T$ are the outputs of a bi-directional RNN.}
\intertext{\textbf{DKF-NL:}}
    q_\phi(z_{1:T}) &= \setN(z_1|\mu_{\phi, 1}(x_{1:T}), \Sigma_{\phi, 1}(x_{1:T})) \\ &\qquad\cdot \prod_{t=2}^T \setN(z_t|g_\phi(z_{1:t-1}, u_{\phi, t}(x_{1:T})), G_{\phi, t}(z_{1:t-1}, u_{\phi, t}(x_{1:T})))
\intertext{
where $g_\phi$ and $G_\phi$ are neural networks, and $\{u_{\phi, t}(x_{1:T})\}_{t=1}^T$ are the outputs of a bi-directional RNN.}
\intertext{\textbf{DKF-CL:}}
    q_\phi(z_{1:T}) &= \setN(z_1|\mu_{\phi, 1}(x_{1:T}), \Sigma_{\phi, 1}(x_{1:T})) \prod_{t=2}^T \setN(z_t|A z_{t-1} + \mu_{\phi, t}(x_{1:T}), \Sigma_{\phi, t}(x_{1:T}))
\end{align*}
where $\{\mu_{\phi, t}(x_{1:T}), \Sigma_{\phi, t}(x_{1:T})\}_{t=1}^T$ are the outputs of a convolutional neural network applied across the time dimension.

The KVAE and CPC models are parametrized in the same way as originally published in \cite{DBLP:conf/nips/FraccaroKPW17} and \cite{infonce}, respectively. In particular, the CPC autoregressive model is parametrized with a forward GRU. The KVAE chain mixture weight RNN is parametrized with a forward LSTM.

\subsection{Auxiliary Generative Models}

The loss used to train the \model auxiliary generative model (\subfigref{fig:applications}{a}) was
\begin{equation*}
    \setL(\psi) = \sum_{n=1}^N \sum_{t=1}^T \ave{\log p_\psi(x_t^n|z_t)}{q^n(z_t)},
\end{equation*}
where $\psi$ parametrizes the generative model:
\begin{equation*}
    p_\psi(x_t^n|z_t) = \setN(x_t^n|f_\varphi(z_t), \sigma^2 I),
\end{equation*}
for $\psi = (\varphi, \sigma^2)$. The loss was estimated using samples $z_t \sim q^n(z_t)$ and optimized using the reparametrization trick to backpropagate through the samples. This is identical to the generative component of the loss used by the SVAE and the DKF.

The loss used to train the CPC auxiliary generative model was
\begin{equation*}
    \setL(\varphi) = \sum_{n=1}^N \sum_{t=1}^T ||f_\varphi(c^n_t) - x^n_t||^2_2,
\end{equation*}
where $c_t$ is the \textit{context vector} output by the recurrent network \citep{infonce} and $f_\varphi$ is the auxiliary generative network.

All generative networks $f_\varphi$ are parametrized as three-layer deconvolutional neural networks followed by a sigmoid function to map to $(0,1)$.

\subsection{Hyperparameters} \label{app:hyperparams}

A learning rate of $10^{-3}$ was used for all runs. All runs swept over recognition covariance parametrized as either diagonal or as a Cholesky factor. The \model was further swept over recognition covariance being data-dependent or data-independent. For all generative baseline methods, the generative covariance was taken to be constant, as is typical in the VAE literature. The generative covariance was parametrized as $\nu^2 I$, with $\nu$ swept from $\{1, 0.1, 0.001\}$.

To enforce stability of the learned latent dynamical systems we required $\rho(A)<1$. A sufficient condition is that the largest singular value of $A$ is less than 1. The set of matrices with singular values less than 1 is convex, simplifying constrained optimization \citep{stable-lds}. We enforce the constraint by clipping the singular values of $A$ to the range $(0,1-\epsilon)$ after each gradient step. We took $\epsilon=10^{-3}$.

The linear and all pendulum experiments were trained for 10K iterations, whereas all tracking experiments were trained for 5K iterations.

The following latent dimensionalities were swept over:
\begin{itemize}
    \item Single-pendulum: $\{4, 8\}$;
    \item Double-pendulum: $\{8, 16\}$;
    \item Tracking: $\{8, 16, 32\}$.
\end{itemize}

For the linear experiments, the latent dimension of all methods was set equal to the problem's true latent dimension.

\cref{tab:lra} shows the optimal parameters for the \model in different experiments.

\begin{table*}[t]
\caption{Optimal hyperparameters for all \model experiments from \cref{sec:experiments}. $\setD_{\mathcal{Z}} = $ \model latent dimensionality; CRC $=$ constant recognition covariance; DRC $=$ diagonal recognition covariance; LR $=$ learning rate.}
\label{tab:hyperparams}
\begin{center}
\begin{small}
\begin{tabular}{lccccccc}
\toprule
Dataset & $\setD_{\mathcal{Z}}$ & CRC & DRC & LR & BS & Train Iters. & Optim. \\
\midrule
Linear & -- & Yes & Yes & 0.001 & 32 & 10K & Adam \\
Pendulum & 8 & Yes & No & 0.001 & 32 & 10K & Adam \\
\texttt{CIFAR10}-Pendulum & 8 & Yes & No & $0.001$ & 32 & 10K & Adam \\
Double-Pendulum & 8 & No & No & $0.001$ & 32 & 10K & Adam \\
\texttt{CIFAR10}-Double-Pendulum & 16 & Yes & No & $0.001$ & 32 & 10K & Adam \\
Tracking & 16 & Yes & No & $0.001$ & 32 & 5K & Adam \\
\texttt{DAVIS}-Tracking & 32 & Yes & No & $0.001$ & 32 & 5K & Adam \\

\bottomrule
\end{tabular}
\end{small}
\end{center}
\label{tab:lra}
\end{table*}

\subsection{Compute Resources}

All experiments except for those on the Tracking and \texttt{DAVIS}-Tracking tasks were run on a single RTX5000 GPU. Tracking experiments were run on a single RTX4090 GPU.

\section{Experimental Details} \label{app:experiments}

Each dataset consists of data sequences of $T=100$ time steps. The number of sequences $N$ in each dataset is as follows:
\begin{itemize}
    \item \textbf{Linear:} $N = 200$;
    \item \textbf{Pendulum:} $N=500$;
    \item \textbf{Double Pendulum:} $N=376$;
    \item \textbf{Tracking}: $N=500$.
\end{itemize}

The \textbf{Pendulum} dataset consists of sequences of 24$\times$24 pixel frames sampled at 0.01s intervals. Pixel intensities are normalized to $(0,1)$ and uncorrelated pixel-level zero-mean Gaussian noise with standard deviation 0.05 is applied to all frames. The \textbf{Double Pendulum} observations are 50$\times$50 pixel frames of a real-life double pendulum system, sampled at 0.0075s intervals. Frames are single-channel grayscale for both tasks.

For the \textbf{Tracking} dataset, the observation frames are 64$\times$64 pixels with three color channels. Each 100 frame-long sequence is equivalent to approximately 5s of real time. The ``head''-direction (represented by an arrow in the observations) is computed using a low-pass filter on the velocity vector.

\section{Further Results} \label{app:further-results}

\begin{figure}[h!]
    \centering
    \includegraphics[width=0.5\linewidth]{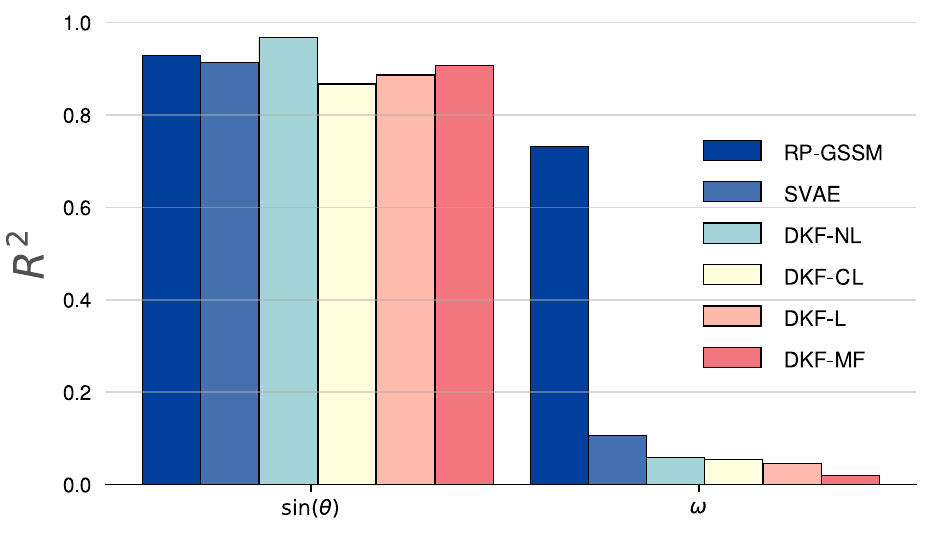}
    \caption{Linear regression \(R^2\) scores to sine of pendulum angle, \(\sin\theta\), and angular velocity, \(\omega\), of all methods trained on the first 64 sequences of the single-pendulum dataset.}
    \label{fig:pendulum-small}
\end{figure}

\subsection{\cref{fig:pendulum-small}}

The \model is the only method tested that is able to infer angular velocity when trained on only 64 data sequences.

\subsection{Additional Regression Results}

\begin{figure}[h!]
    \centering
    \includegraphics[width=\textwidth]{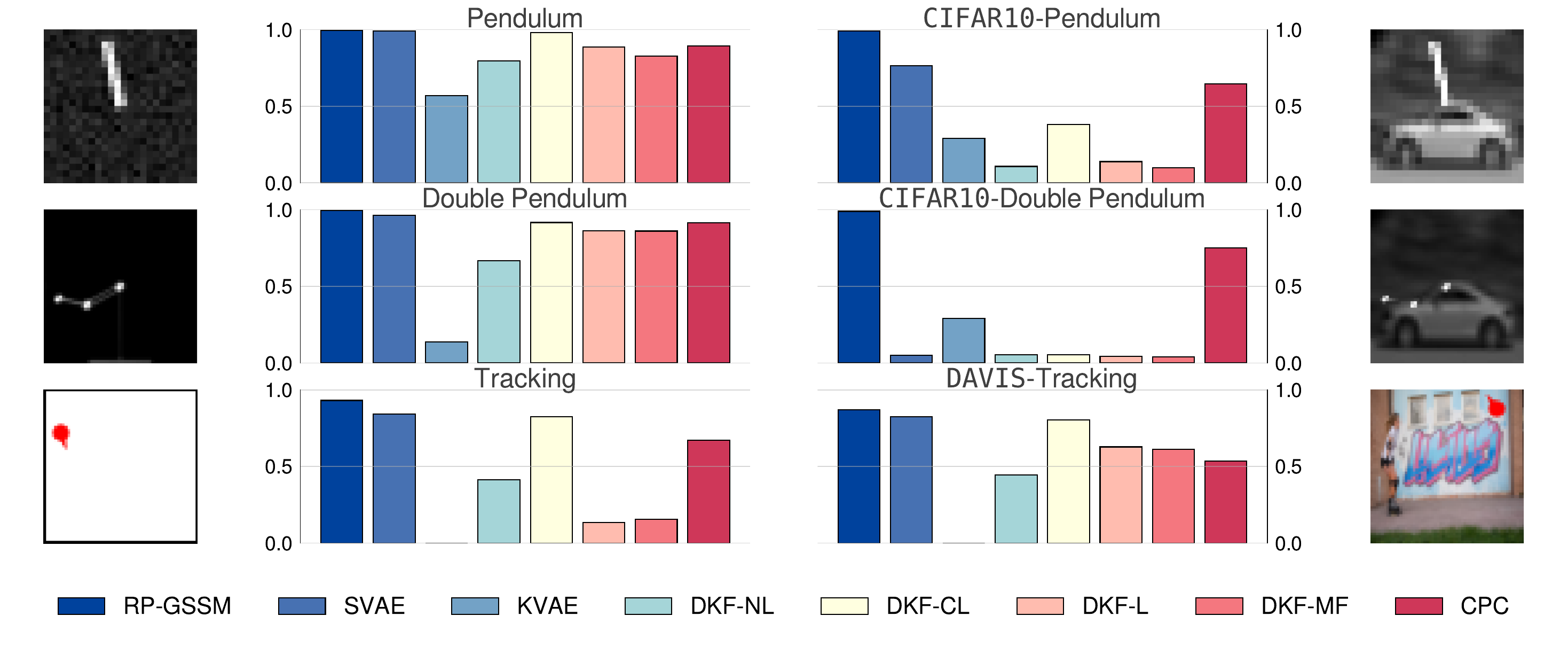}
    \caption{Mean kernel ridge regression $R^2$ scores.}
    \label{fig:all-r2s-kernel}
\end{figure}

\cref{fig:all-r2s-kernel} is identical to \cref{fig:r2s}, but using kernel ridge regression instead of linear regression. Regression was performed with an RBF kernel and with a fixed regularization constant across all models and tasks.

\begin{figure}[h!]
    \centering
    \includegraphics[width=\textwidth]{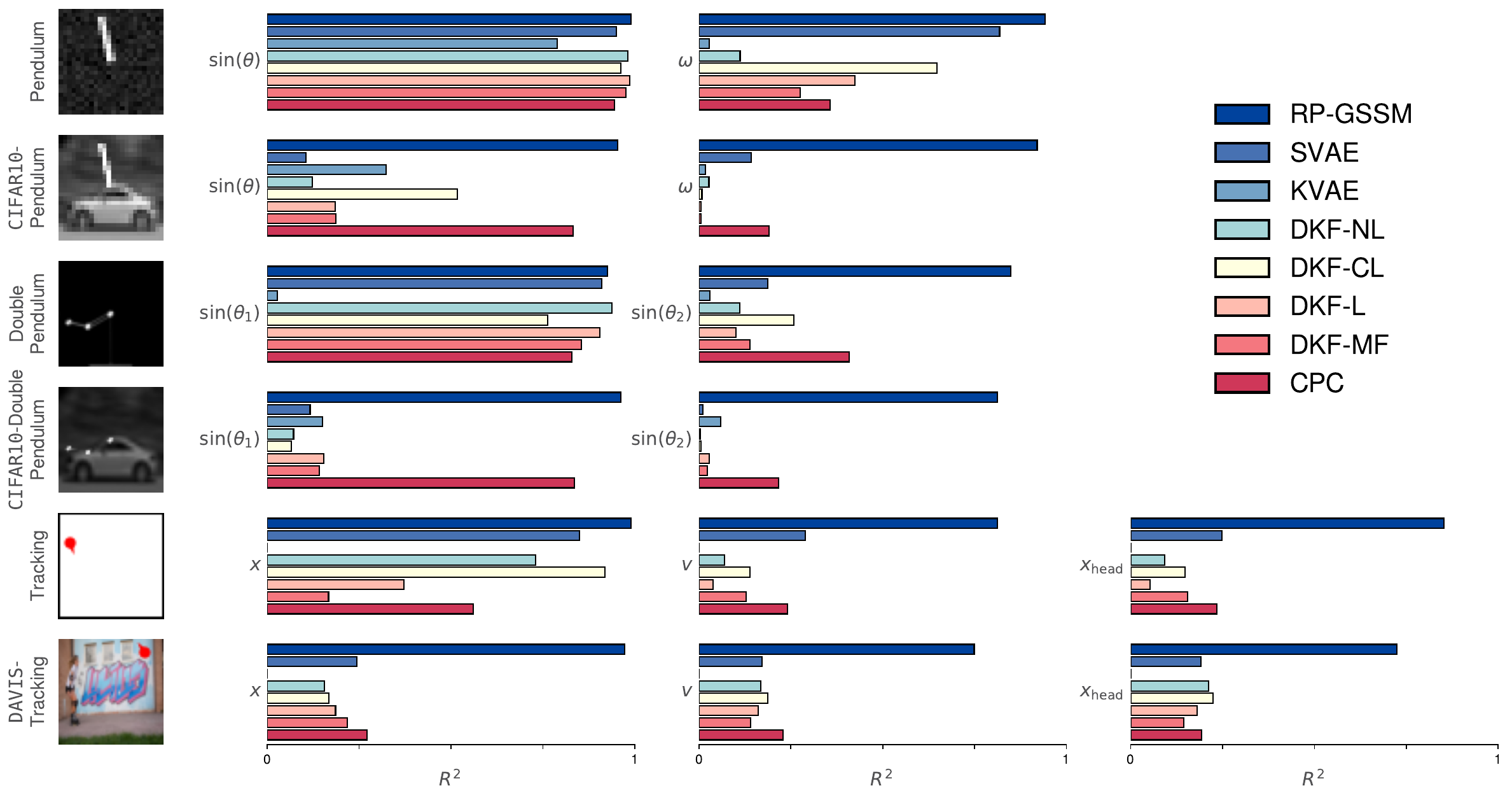}
    \caption{Linear regression $R^2$ scores to all target variables.}
    \label{fig:complete-r2s-linear}
\end{figure}

\begin{figure}[h!]
    \centering
    \includegraphics[width=\textwidth]{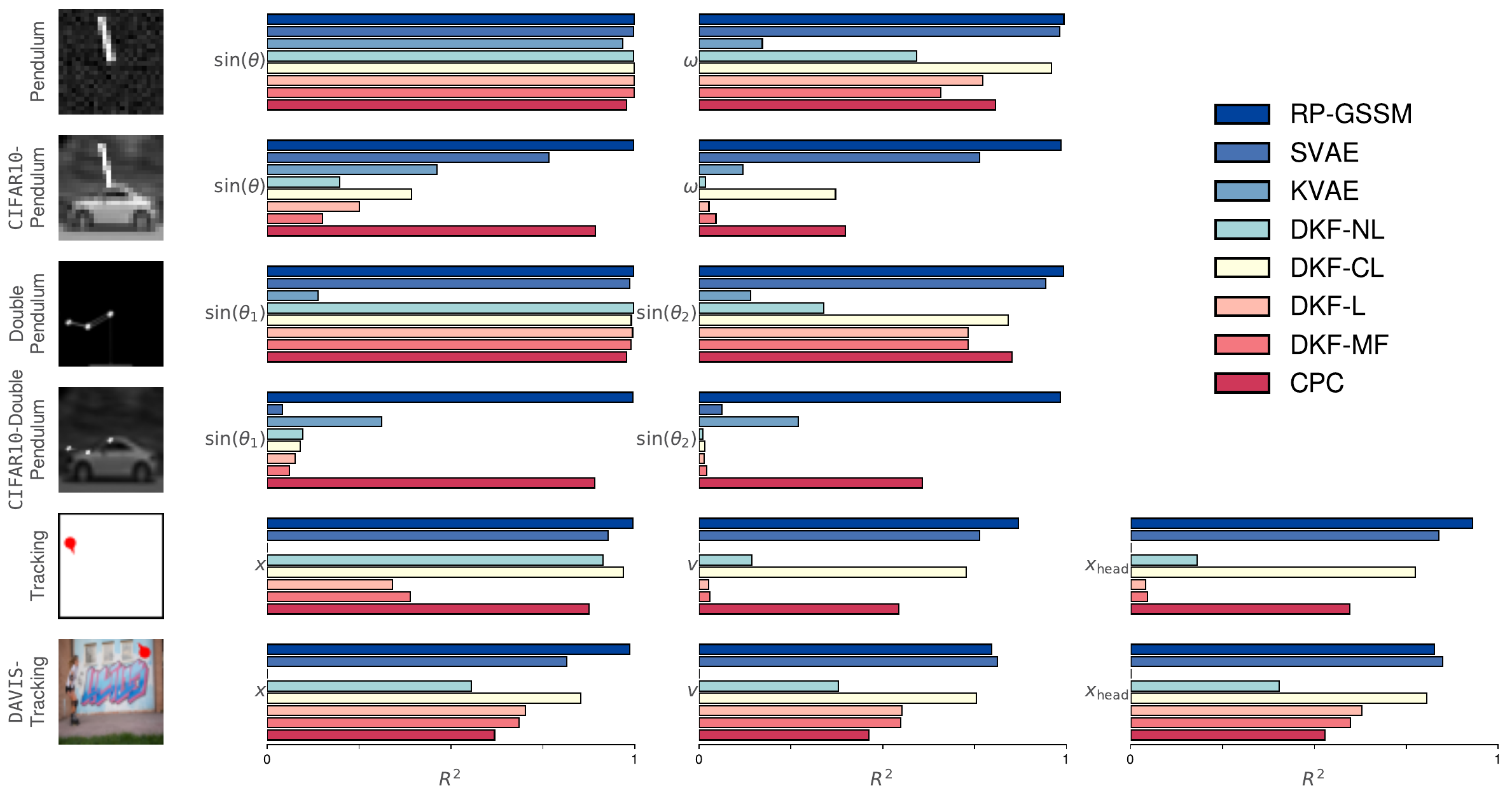}
    \caption{Kernel ridge regression $R^2$ scores to all target variables.}
    \label{fig:complete-r2s-kernel}
\end{figure}

\cref{fig:complete-r2s-linear,fig:complete-r2s-kernel} show $R^2$ scores from linear and kernel ridge regression, respectively, to all target variables. The target variables are as follows:
\begin{itemize}
    \item \textbf{Top two rows:} sine of angle, $\sin\theta$, and angular velocity, $\omega$;
    \item \textbf{Middle two rows:} sine of primary angle with respect to the origin, $\sin\theta_1$, and sine of secondary angle with respect to the primary angle, $\sin\theta_2$;
    \item \textbf{Bottom two rows:} rat position, $x$, rat velocity, $v$, and rat head direction, $x_{\mathrm{head}}$. These have dimensionality 2, 2, and 1, respectively.
\end{itemize}

\subsection{Computational Complexity and Runtimes}

\begin{figure}[h!]
    \centering
    \includegraphics[width=\textwidth]{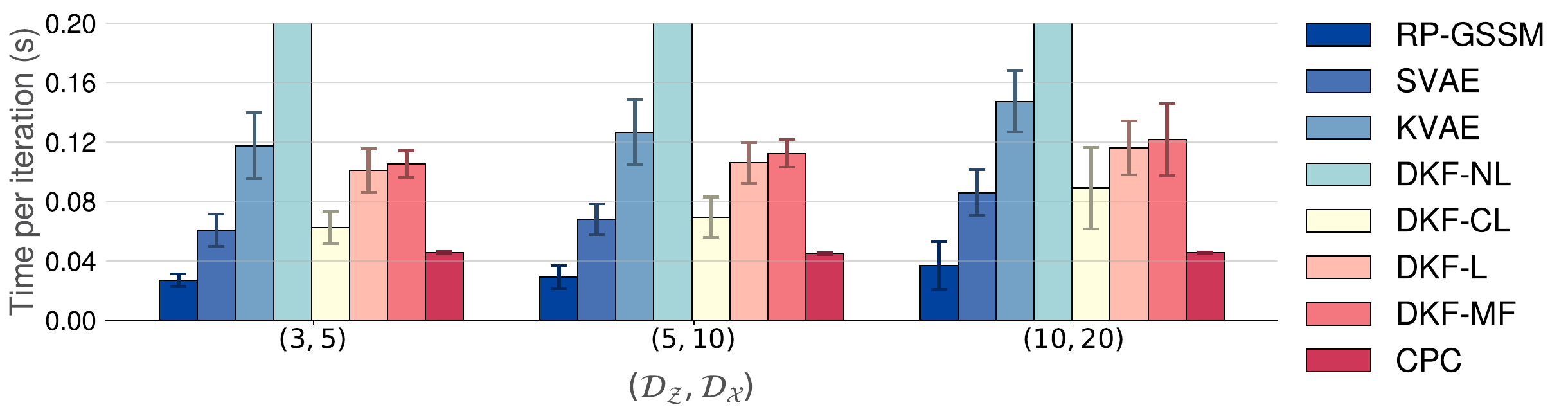}
    \caption{Average times per iteration across the full sweeps for the linear tasks. Error bars indicate standard deviation. The times for DKF-NL (approximately 0.3s) are significantly higher than for other models, so they are cropped for visual clarity.}
    \label{fig:times}
\end{figure}

\begin{figure}[h!]
    \centering
    \includegraphics[width=\textwidth]{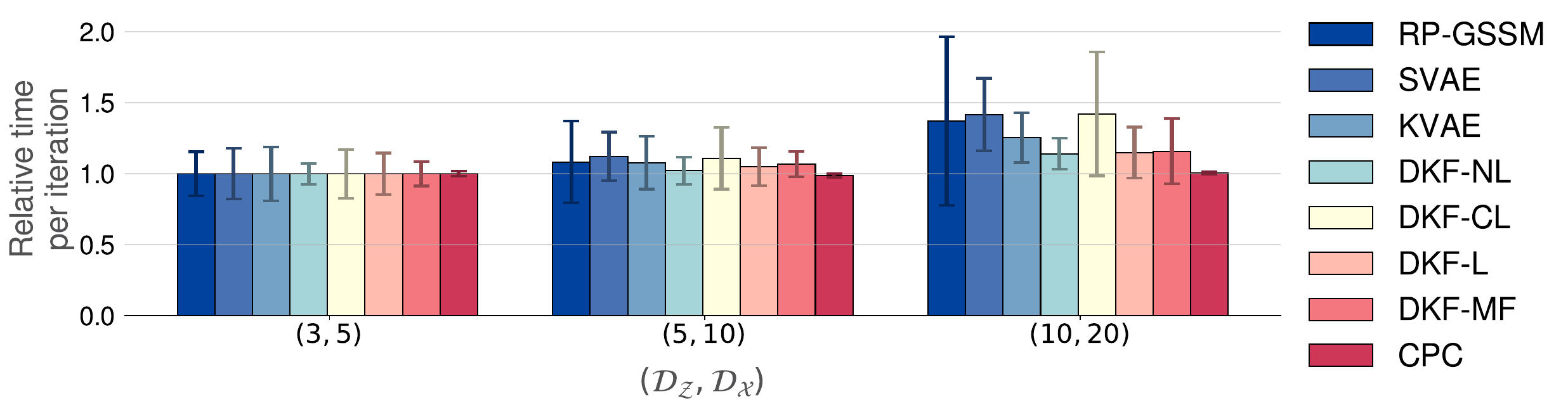}
    \caption{Average times per iteration, normalized relative to the mean time per iteration for the smallest linear task, across the full sweeps for the linear tasks. Error bars indicate standard deviation.}
    \label{fig:relative-times}
\end{figure}

\cref{fig:times} shows the mean time it takes for each model to complete one iteration, averaged over all runs in the sweeps, for the three linear tasks. The \model has the fastest iterations. However, these times are implementation-dependent and may not necessarily be representative of true computational complexity. For completeness, we normalize each model's times relative to its mean time on the smallest linear task. Normalized mean times are shown in \cref{fig:relative-times}; the \model times scale with latent and observation dimensionalities at a similar rate to the baselines models.

\section{Licenses} \label{app:licenses}
We release the \model model and code under the CC BY-NC 4.0 license.

The assets used in this work have the following licenses:
\begin{itemize}
    \item The \texttt{CIFAR10} dataset: MIT license;
    \item The \texttt{DAVIS} dataset: BSD license;
    \item The code for the SVAE and DKF models \citep{zhao2023revisiting}: MIT license;
    \item The KVAE model: unknown license;
    \item The CPC model: unkown license.
\end{itemize}

\end{document}